\documentclass[12pt,a4paper]{article}
\usepackage[english]{babel}
\usepackage{graphicx}

\newcommand{\alt}{\mathbin{\lower 3pt\hbox
   {$\rlap{\raise 5pt\hbox{$\char'074$}}\mathchar"7218$}}}
\newcommand{\agt}{\mathbin{\lower 3pt\hbox
   {$\rlap{\raise 5pt\hbox{$\char'076$}}\mathchar"7218$}}}

\begin{document}
\setcounter{footnote}{0}
\setcounter{equation}{0}
\setcounter{figure}{0}
\setcounter{table}{0}
\vspace*{5mm}

\begin{center}
{\large\bf How to realize "a sense of humour" in computers ? }

\vspace{4mm}
I. M. Suslov \\
P.L.Kapitza Institute for Physical Problems,
\\
119337 Moscow, Russia \\
 E-mail: suslov@kapitza.ras.ru
\vspace{4mm}
\end{center}

\begin{center}
\begin{minipage}{135mm}
{\large\bf Abstract } \\
Computer model of a "sense of humour" suggested
previously [1\,--\,3] is raised to the level of a realistic
algorithm.

\end{minipage}
\end{center}
 \vspace{5mm}


\begin{center}
{\bf 1. Introduction}
\end{center}
\vspace{3mm}

In the previous papers of the present author \cite{1,2,3},
the general scheme of information processing was suggested,
which naturally leads to a possible realization in computers
of the simplest human emotions and, in particular,
a "sense of humour". The aim of the present paper is to develop
this general scheme to a level of a realistic
algorithm.\,\footnote{\,At this work, we have in mind the
simplest samples of humour related with the primary processing of
information. The higher levels of information processing can be
treated similarly \cite{3} but require more complicated
constructions. }

Briefly, the previously formulated model \cite{1} consists in the
following. Let a succession of symbols (or "words")  $A_1$, $A_2$,
$A_3\,\ldots$ is entering the input of a processor. Each word
$A_n$ is associated with a set of images $\left\{ B_n \right\}$.
The problem consists in the choice from each set
$\left\{ B_n \right\}$ of the single image $B_n^{i_n} $, which
is implied in a given context. We consider that the text
is "understood", if the succession of images
$B_1^{i_1}$, $B_2^{i_2}$, $B_3^{i_3}\,\ldots $ is put in
correspondence to the sequence of symbols
$A_1$, $A_2$, $A_3\,\ldots$; the former can be considered
as a certain "trajectory" (Fig.\,1).
\begin{figure}
\centerline{\includegraphics[width=5.1 in]{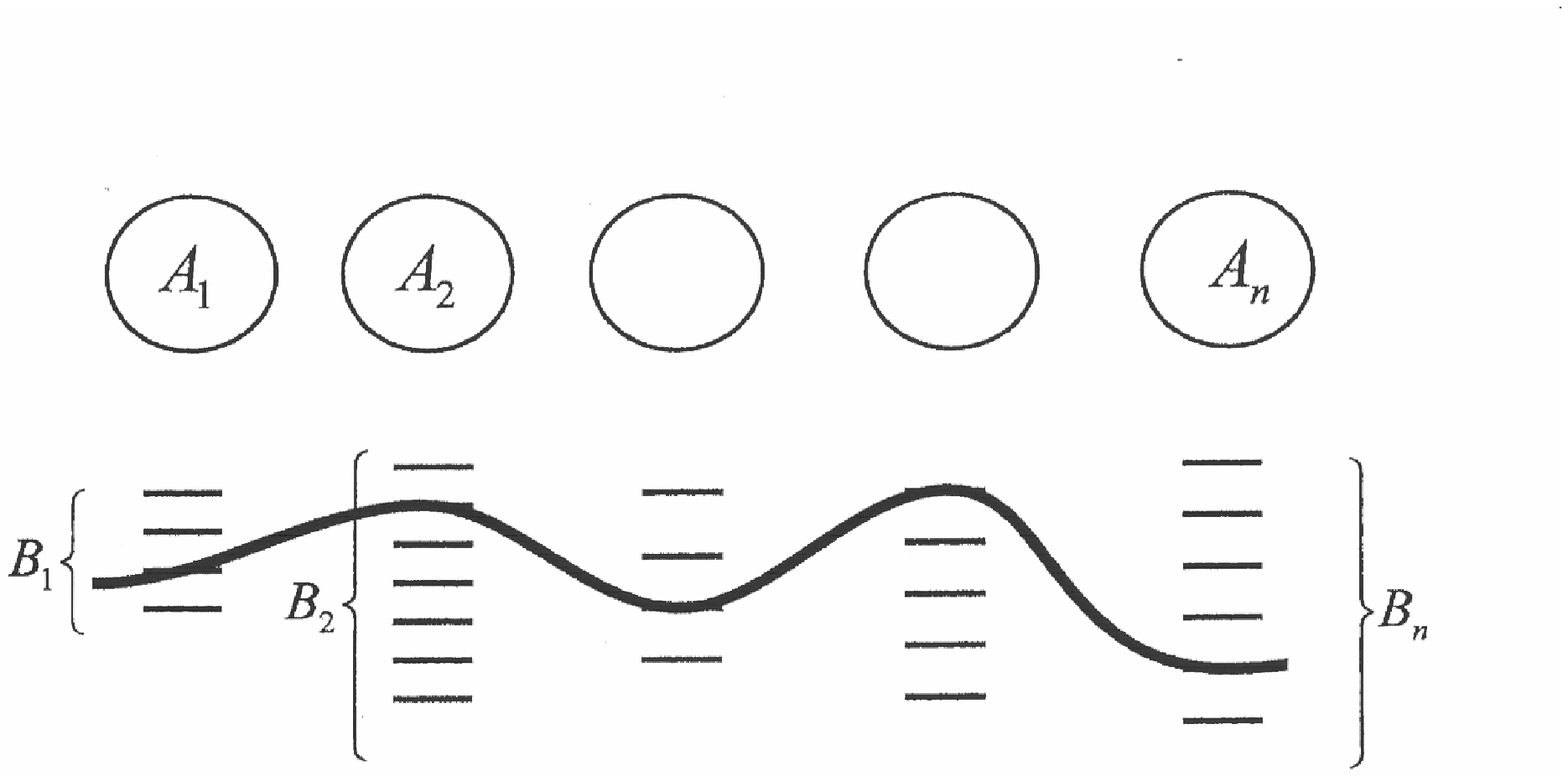}}
\caption{The scheme of information processing:
each symbol $A_n$ is associated with a set of
images $\left\{ B_n \right\}$, from which a single image
$B_n^{i_n} $ should be chosen; succession
$B_1^{i_1}$, $B_2^{i_2}$, $B_3^{i_3}\,\ldots $ can be
considered as a certain "trajectory". } \label{fig1}
\end{figure}

In principle, the algorithm consists in the following:
(a) all possible trajectories are composed; (b) a certain
probability is ascribed to each trajectory;  (c) the most probable
trajectory is chosen. Only step (b) is nontrivial, i.e.
the algorithm for estimation of the probability for a given
trajectory. Such algorithm should be based on the correlation
of images, which can be studied in the process of "learning"
on a sufficiently long "deciphered" text, i.e. the text
written in images and not words.

Any specific realization of such algorithm needs a number of
operations, expo- nentially growing with the length of the text;
so the algorithm is able to treat the text, which contains not
more than a certain number ($L$) of symbols. To deal with
longer texts, one can suggest  the following procedure.
During  processing of the first $L$ words, one
remembers not one but several ($M$) most probable trajectories.
As the next step, the fragment of the text between the second and
$(L+1)$-th word is considered, and all possible continuations are
composed for each of the $M$ trajectories. Then again $M$
most probable of them is remembered, and so on.
In general, the process looks in the following manner (Fig.2,a):
\begin{figure}
\centerline{\includegraphics[width=5.1 in]{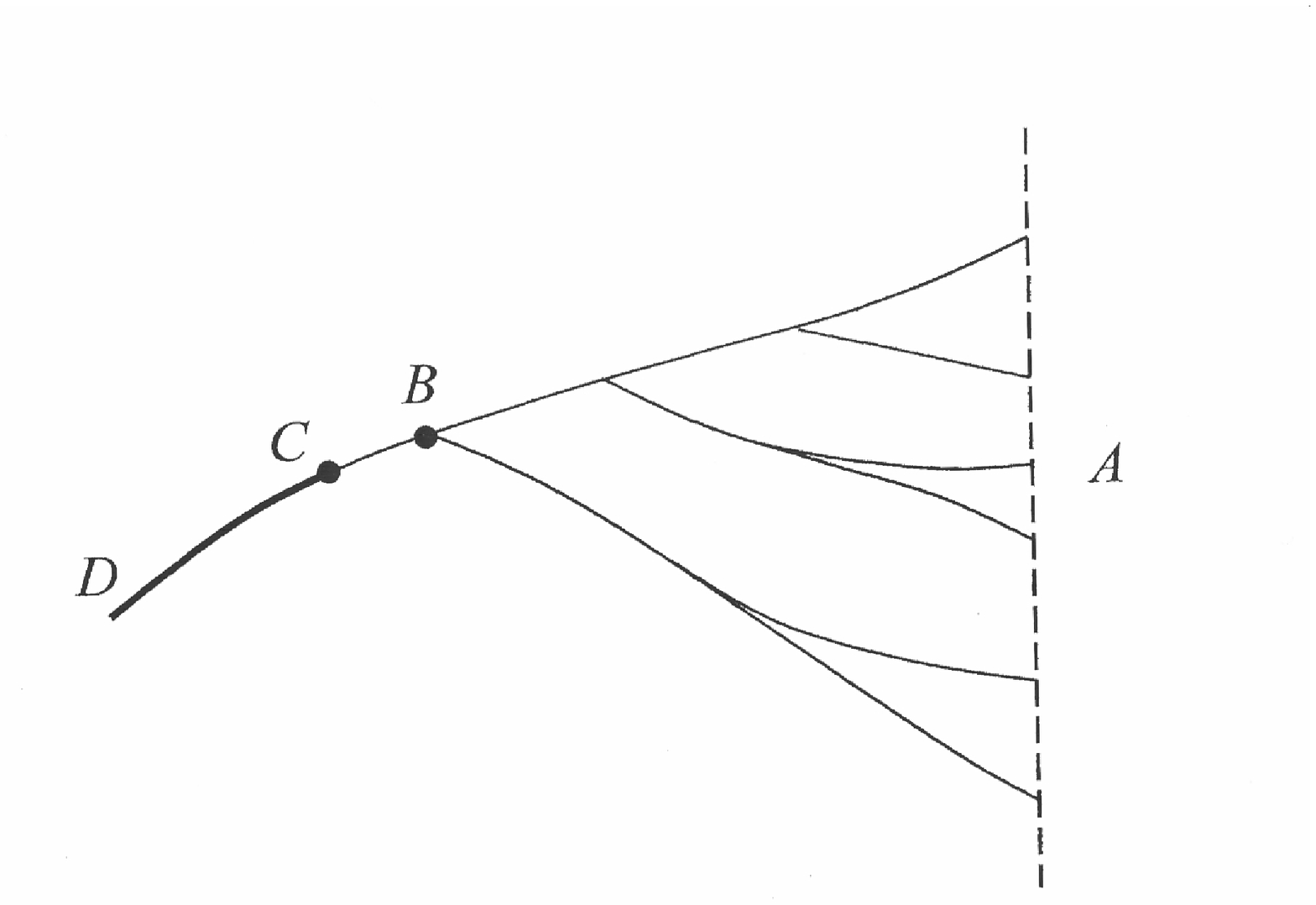}}
\caption{A visual imagination of  information
processing: thin lines are the trajectories contained in
the operative memory (or subconsciessness),  $A$ is a front edge,
$B$ is the point where branching is over,  $CD$ is a portion
of a trajectory transmitted to the output of the processor
(or consciousness).
 } \label{fig2}
\end{figure}
the trajectory is branched strongly near its front edge $A$,
the branching is ended in a certain point $B$, and the deciphered
part $CD$ is transmitted to the output of a processor (with a
certain delay $AC$).

At first sight,  point $C$ should always go behind  point
$B$ or coincide with it. However, for a biological system a delay
$AC$ should have upper bound on the time scale, since
information processing is carried out in subconsciousness and no
information appears in consciousness before a deciphered
trajectory $CD$ has reached it.
If a distance $AB$ is
sufficiently large,  point $C$ begins to outrun  point $B$
\cite{1},  i.e.  the most probable trajectory is transmitted to
consciousness, though the competing versions are still conserved
in the operative memory.  If, in the following, the probability
of the transmitted trajectory becomes lesser than for one of
competing versions, then a characteristic malfunction
occurs, which can be identified with "a humorous effect" on
the psychological grounds.

It is easy to see, that to endow a computer by a "sense of
humour" one should be able to solve a "linguistic problem",
i.e. recognition of a succession of polysemantic images,
which also arises in the machine translation researches
\cite{4,5}\,\footnote{\,In spite to similarity with the problem
of machine translation, our aim is somewhat different; this
difference is the main subject of subsequent discussion.
Usual strategy of machine translation contains three stages
(analysis of the source text, transfer to another language, and
generation of the target text); we are interested only in the
first of them but developed as far as possible.  In existing
programs this stage is not very advanced (see Fig.\,6.1 in
\cite{4}). }. This general problem can be divided into several
more specific problems:

(1) one should compose a list of images, which are actual
for a population;

(2) each word of a given language should be associated
with a certain set of images;

(3) one needs a sufficiently long text for learning,
i.e. the text written in images and not in words;

(4) one should formulate the educational algorithm,
reducing to the study of correlations between images;

(5) on the basis of this algorithm, a rule for estimating
probability of a finite succession of images should be
worked out.

In principle, problems  1\,--\,3 are trivial, but they need
enormous amount of qualified work; it is difficult to imagine
that such work can be made specially for realization of a "sense
of humour". Below we consider the possibility to solve these
problems in the automatic regime, having in mind the linguistic
information, which is gathered
in the systems like  ABBYY
Lingvo.  Problems 4,\,5 are more complicated: they cannot be
solved pure theoretically and require long period of
experimentation by trial and error method. We suggest here only a
preliminary variant of their solution, having in mind to
emphasize some essential points. The practical algorithm can be
formulated on the basis of approaches worked out in the
field of machine translation \cite{4,5}.

\vspace{6mm}
\begin{center}
{\bf 2. Ideal language as a limiting case.}
\end{center}
\vspace{3mm}

Suppose that there exists some language (let
it be Latin, for definiteness), which in a certain approximation
can be considered as ideal. We define the ideal language as
a language, whose words are in one to one correspondence
with images.

Then  problems 1\,--\,3 are solved trivially. To compose a
list of actual images (problem 1), it is sufficient to
write down all Latin words; they can be numerated in
alphabetical order. If we want to recognize texts in English,
then problem 2 is solved with a help of the English\,--\,Latin
dictionary: it is sufficient to write down all variants for
a translation of a given English word to Latin. As a text for
learning (problem 3), one can take any literary Latin text.
Now let us discuss problems 4,\,5.

\vspace{2mm}

{\it Learning algorithm.}  A simplest algorithm for learning
consists in construction of the correlation matrix
$A_{ij}$ where indices $i$ and $j$ run all images in
alphabetical order.  We accept that $A_{ij}\equiv 0$ before
education. One step of education consists in the analysis of a
separate sentence. Each sentence by definition expresses a
closed thought, and hence all words in it are inter-related
(words are equivalent to images in the ideal language). We
increase by unity an element $A_{ij}$ of the correlation matrix
for any pair $(i,j)$ of words entering this sentence (Fig.\,3),
\begin{figure}
\centerline{\includegraphics[width=5.1 in]{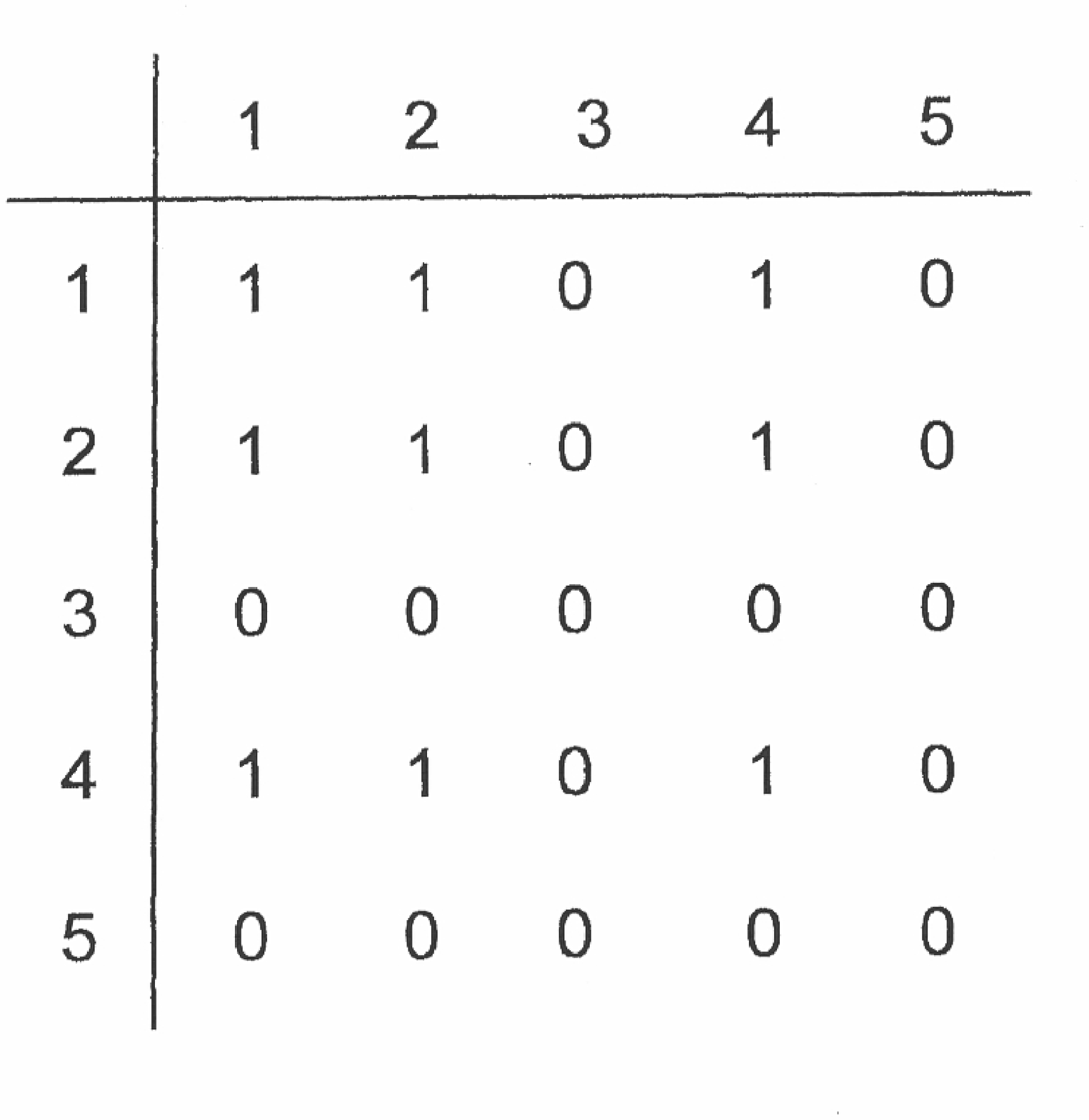}}
\caption{A change of the correlation matrix $A_{ij}$ in the
result of processing the sentence 4--2--1.
 } \label{fig3}
\end{figure}
$$
\Delta A_{ij}=1 \,,
\eqno(1)
$$
including a case $i=j$ (see below). Of course, such learning rule
leads to inevitable errors. Indeed, syntactic bonds
in the sentence have a tree character (Fig.4), and the existence
\begin{figure}
\centerline{\includegraphics[width=5.1 in]{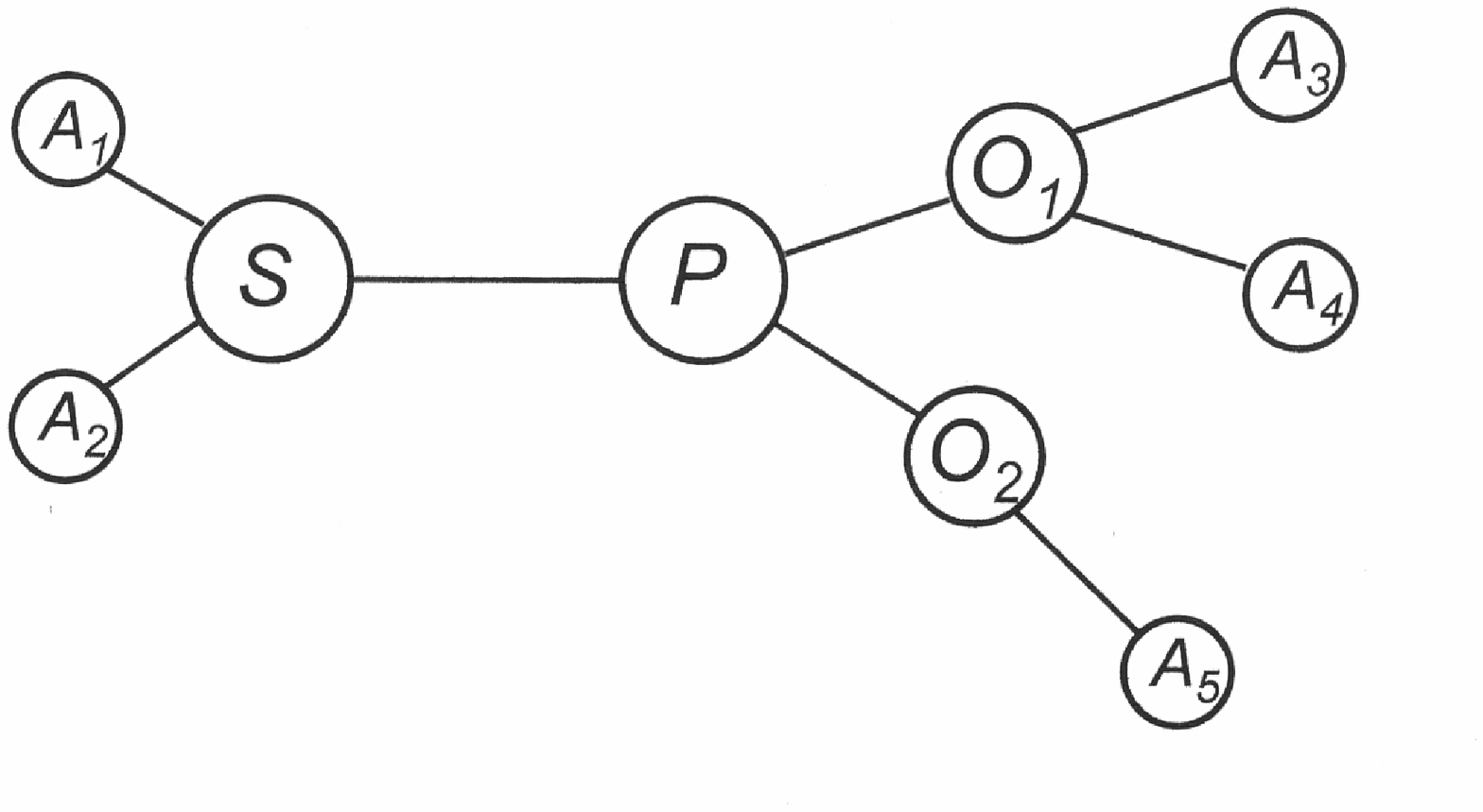}}
\caption{ Syntactic bonds in the sentence have a tree strusture:
a Subject ($S$) is related with a Predicate ($P$), which can be
related with Objects ($O_1$, $O_2$);
the former and the latter can have Attributes
($A_1$\,--\,$A_5$).} \label{fig4}
\end{figure}
of associative links can be guaranteed only for syntactically
connected images.  However, absolutely uncorrelated images can
appear in one sentence only with probability $\sim\left(n/N
\right)^2$ ($n$ is the number of words in a sentence, $N$ is the
number of words in a language), which should be compared with
probability $\sim k\left(n/N \right)$ for associatively related
images (\,$k$ is the correlation coefficient). Evidently, such
procedure allows to reveal correlations effectively for
a wide range of $k$ values  ($1\agt k \agt 10^{-3} \div
10^{-4}$).

There is practically no alternative to this algorithm,
since the syntactic analysis can be made by a computer
only with large percentage of errors \cite{4}. On the other hand,
an educational algorithm based only on syntactic connections
is also not very good: a sentence may contain description of
oft-repeating situations (e.g. {\it
"A herdman
drives a herd"}) where all images are associatively related. As the last
argument, we can note that the syntactic analysis plays no
essential role in human learning: the children and
poorly-educated people are sufficiently good in conversational
language, though they have no idea on syntax.

It is evident from the given estimates, that short sentences
are more preferable for learning: the error is lesser for small
$n$, and the whole process is more effective. Indeed, treatment
of two sentences consisting of 10 words requires consideration of
$10^2\cdot 2=200$  pair bonds (most of which are ineffective),
while the analysis of one sentence containing 20 words deals with
$20^2=400$ pair bonds. As for avoiding long sentences, it is not
related with the essential waste of time or resources.

\vspace{2mm}

{\it Probability of trajectory.} When the correlation matrix
$A_{ij}$ is formed in the result of learning on a
sufficiently long text, the probability of a finite
succession of images can be defined
as\,\footnote{\,Normalization of probability remains
arbitrary, but it is not essential for comparison of
trajectories.}
$$
p=\sum\limits_{i,j\,\,(i\ne j)} A_{ij} \, \qquad
\eqno(2)
$$
where $i$ and $j$ run the images contained in this succession.
The algorithm is based on the pure
analog principle, so the combinations of images appear
to be more probable if they  frequently occur in the
learning text. We have excluded the terms with $i=j$ from the
sum in Eq.2, since the probability we are interested in should
characterize the degree of connectedness of a given trajectory.

At first sight, the analog character of the algorithm
requires to introduce the condition $i\ne j$ also in the
learning rule (1).  In fact, it is not so, since
learning and recognition are carried out in somewhat different
conditions: we use only closed sentences for learning, while
any fragment of text can be given for recognition,
independently on the bounds of sentences.  If the condition
$i\ne j$ is introduced in (1), then self-correlation of images
appear to be practically zero: repeated use of the same word
is usually considered as a stylistic mistake\,\footnote{\,If it
is not an article or a technical word.} and practically never
occurs in the literary text.  It is evident from the general
principles, that correlation of an image with itself should be
maximal: it is provided by inclusion of the case $i=j$ in the
learning rule (1). In the course of recognition,
self-correlation of images plays essential role:  the connected
fragment of the text contains usually some kind of the "main
hero", whose image is present in almost any sentence. As a
result, existence of the same image in two neighbouring sentences
is rather typical, and its self-correlation is important for
an adequate estimation of the connectedness of the text.

\vspace{6mm}
\begin{center}
{\bf 3. Real language: nobody wanted "as worse".}
\end{center} \vspace{3mm}

Of course, any real language is very far from ideal:
almost any word is associated with a lot of images, while
any image can be described by different words. It looks, as if
some evil spirit interfered in human life and spoiled specially
all existing languages\,\footnote{\,Legend on the Babilon Towel
is probably arised from such impression.}.
In fact, nobody wanted "as worse" and nobody was specially
entangling the situation: ambiguity of real languages is a
natural consequence of their evolution.

The first words of the ancient man were formed according to a
principle "I sing what I see"\,\footnote{\,Tradition
of some Asia tribes.}:  e.g. words {\it "fox"},
{\it "wolf"}, {\it "bear"}  arised from the shouts of
hunters warning about appearance of the
corresponding animal. These words had  clear associations and
did not possess any ambiguity (Fig.\,5,a). When inter-relations
\begin{figure}
\centerline{\includegraphics[width=5.1 in]{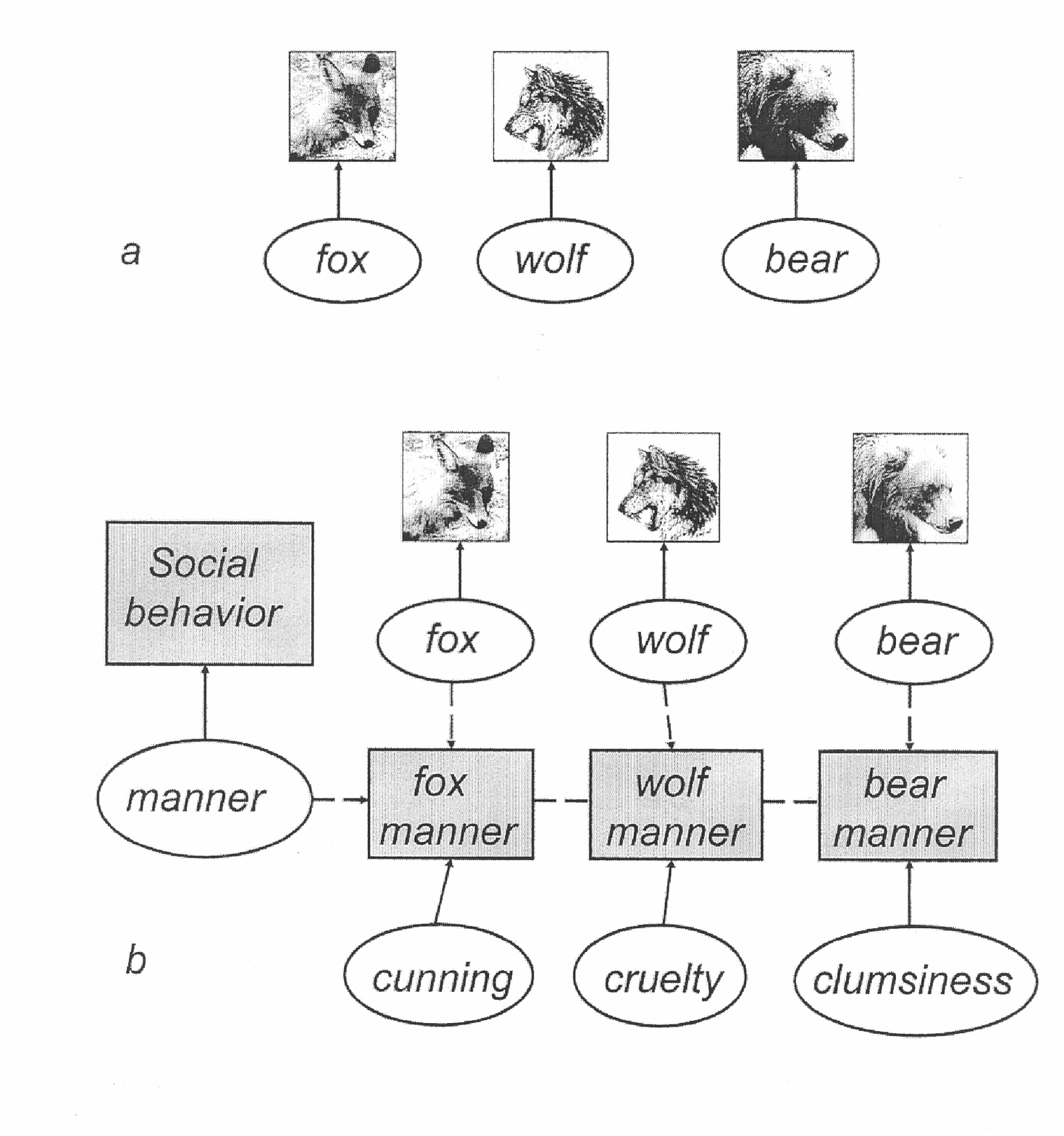}}
\caption{(a) The first stage in evolution of  language: words
and images are in one to one correspondence. (b) The next stage
of evolution: the main mechanism of arising ambiguity is
shown. For clarity, images are given in square frames.} \label{fig5}
\end{figure}
between people became more complicated, some interest arised to
the problems of social behavior: the words like {\it
"manner"}, {\it "character"} came to life. Combination of these
words with already existing   led to appearance of complex
images:   {\it "fox manner"}, {\it "wolf manner"}, {\it
"bear manner"}; these notions appeared to be useful and was
denoted by special words: {\it "cunning"}, {\it "cruelty"},
{\it "clumsiness"}  (Fig.\,5,b). One can see, that the
language adequately reacted on the change of the
situation: arised new images gave birth to new words, and the
total number of words remained in correspondence with the number
of images (Fig.\,5,b). The entanglement of  language appears
already at this primitive stage: on one hand, the
synonym ambiguity arises (one can say {\it "cunning"} or {\it "fox
manner"}), on the other hand, old words acquire new meanings (the
word {\it "fox"} now denotes not only "a red thing with a big
tail" but also "cunning aunt Mary").  We see, that ambiguity
of  language is inavoidable consequence of its development:
initially new images are explained by old words,
but later on the special names are invented for them; however,
associative relations with the old words nobody is
able to abolish.

Fortunately, the entanglement of language
related with its
development is easily removable. It is possible to distinguish
the main meaning for each word (solid arrows in Fig.5,b and
below) and its secondary meanings (dotted arrows). If each word
is ascribed to an image, associated with its main
meaning, then one to one correspondence between words and images
is restored (Fig.\,5,b).

Unfortunately, there are another reasons for arising
ambiguity, which are external from viewpoint of language.
If in two provinces the same image was named
by the different words, then unification of these provinces in one
state makes both words to be admissible. As a result, irreducible
synonyms arise (Fig.\,6,a), under which we understand
\begin{figure}
\centerline{\includegraphics[width=5.1 in]{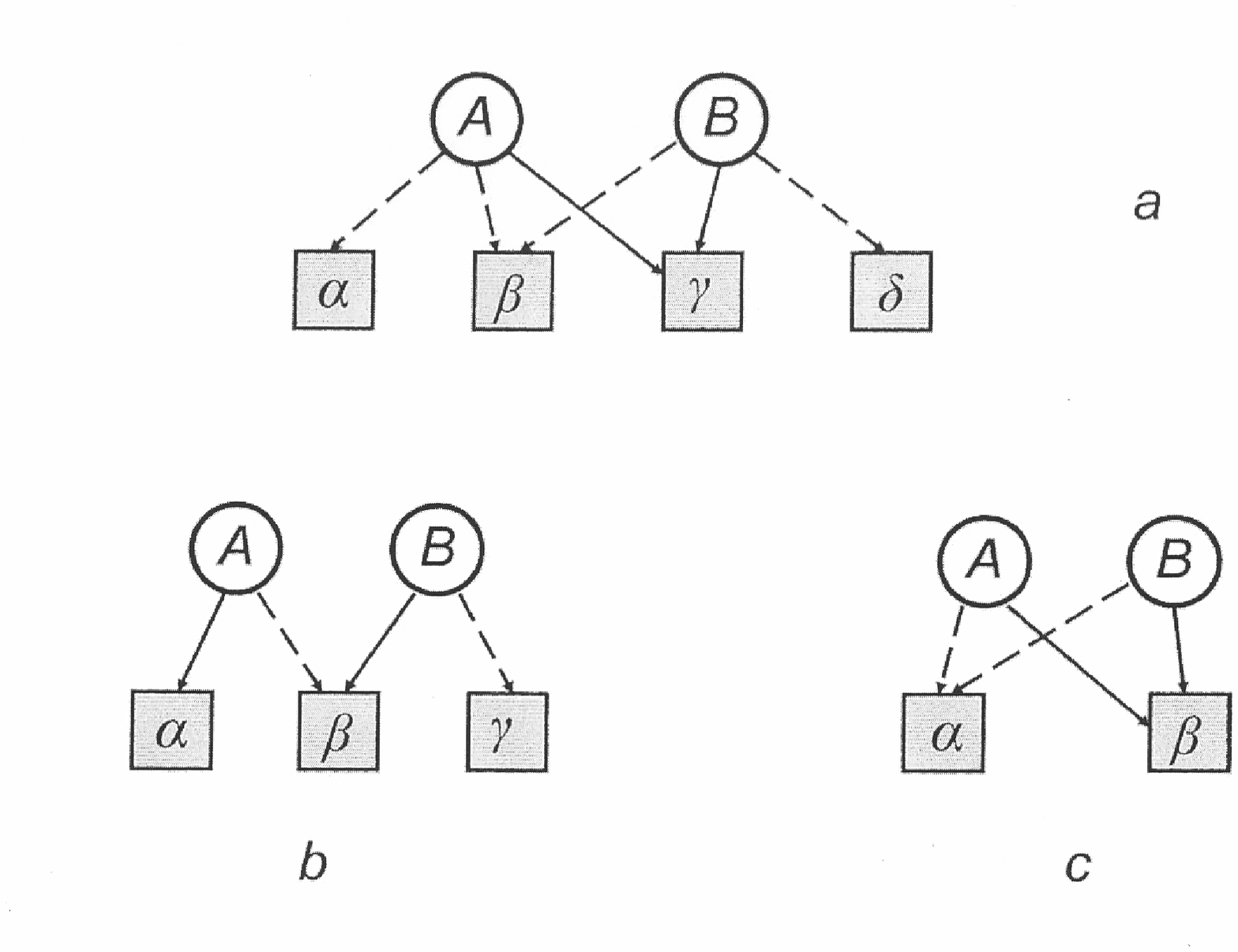}}
\caption{(a) Words $A$ and $B$ are irreducible synonyms, since
their main meanings correspond to the same image $\gamma$.
(b) Words  $A$ and $B$  are synonyms in respect to the image
$\beta$, but they are reducible, since their main meanings are
different.   (c) The perfect synonyms have coinciding both
main and secondary meanings. } \label{fig6}
\end{figure}
the words with the same main meaning; they are opposite
to reducible synonyms
(Fig.\,6,b), whose main meanings are different. The same effect
is produced by the social segregation of society: for example,
the sexual objects are associated with the sleng words by
poorly educated people, while the neutral names are given to
them in the aristocratic
circles, and the scientific terms of the Latin origin are
suggested by a medical society.  The same mechanism can work in
the opposite direction:  if one word was used as a name for
completely different images  (e.g. word {\it "like"} in English),
then homonyms arise, i.e. the words with several main meanings.

Irreducible synonyms and homonyms are in fact the defects of
language: they arise by occasional reasons and
there is no convincing motivation for their existence.

\vspace{6mm}
\begin{center}
{\bf 4. Real language instead of ideal.}
\end{center}
\vspace{3mm}

The analysis given in Sec.3 allows to understand, what kind
of corrections should be introduced in the algorithm, if
we want to use the real language (let it be German)
instead of ideal (Latin in Sec.2).

In order to obtain the list of images, it is sufficient
(in the first approximation) to write down all German words
and associate them with their main meanings (i.e. a word is
considered as a symbolic name for an image, which corresponds
to its main meaning). In fact, the whole algorithm of Sec.2
remains unchanged in the first approximation: a set of images
associated with an English word is obtained with the help of the
English\,--\,German dictionary, the learning is carried out on
the German texts, etc. The main error, intrinsic for such
procedure, consists in the fact that correlations between
images are replaced by the  correlations
between German words.  However, it looks possible to ignore
this error,  because a man
acts in the same manner.
In human practice it is not customary  to
compose the lists of images, while the lists of words
(dictionaries) are known to everybody.
No long texts written in images are available, though
long texts written in
words (books) are well-known.  As a result,
replacement of images by the corresponding words becomes
inevitable in human education.  Comparatively small error of such
education is related with the following:

(a) For most words, their "main meanings" are indeed main, i.e.
words are used in this sense with overwhelming probability;

(b) The secondary meanings of a word are logically
related with the main one (see Fig.\,5,b), and correlations of
the whole conglomerate of meanings  with other  words are close to
correlations for the main meaning;

(c) It is customary in human practice, that learning is carried
out on a standard set of "classical" texts. It is assumed
that the "classic" writers express their thoughts more clearly
in comparison with other people, and the implied image is usually
denoted by the word, which has this image as a main meaning.
In any case, learning on the same texts leads to the same
education for different people; so the people are able to
understand each other, even if such education is far from ideal.

Let us discuss now, what kind of corrections can be really made to
take into account non-ideality of language.
\vspace{2mm}

{\it Irreducible synonyms.} Existence of irreducible synonyms
is displayed in the fact that in our list of images some of
them will be repeated by several times.
To regulate a situation,
it is convenient to accept a viewpoint that the perfect
synonyms (Fig.\,6,c) practically do not exist: if, in some moment,
words $A$ and $B$  were equivalent (coinciding both in the main
and secondary meanings), then their equivalence is spoiled with a
flow of time: they are differently overgrown by new meanings
(Fig.\,6,a) and begin to use in different
contexts.\,\footnote{\, For a situation in Fig.\,6,a:
if both images $\alpha$ and $\gamma$ can appear in some context,
then image $\gamma$ should be denoted by word $B$, while the use
of word $A$ leads to real ambiguities.}
From this point of view, irreducible synonyms mark small
variations of the main meaning and correspond to close but
different images.

Correlations of these slightly different objects with other
images are described adequately by the matrix $A_{ij}$
obtained in accordance with the learning rule (1) (since in
fact correlations are studied between words). Some problems
arise in respect to
correlations between synonyms themselves; they are analogous to
the problems related with self-correlation of images (Sec.\,2).
The use of two synonyms in one short sentence is not desirable in
the same sense, as a repeated use of the same word. As a result,
the learning procedure of Sec.\,2  will lead to the practical
absense of correlations between close images, while such
correlation is large from the general principles.

As a model for irreducible synonyms one can accept that appearing
image is denoted by word $A$ with probability $p_A$,
by word $B$ with probability $p_B$, by word $C$ with probability
$p_C$, etc.\,\footnote{\,Such model implies that the difference
between close images has a symbolical character and
no attention is given to it in human practice. }
Then it is easy to show (see Appendix) that the block of the
correlation matrix, corresponding to synonyms $A$, $B$,
$C\,\ldots$, should have a following structure
$$
\left ( \begin{array}{cccc} S_{AA}    &
\sqrt{S_{AA} S_{BB}}  & \sqrt{S_{AA} S_{CC}}& \ldots \\
\sqrt{S_{BB} S_{AA}} & S_{BB}     &  \sqrt{S_{BB} S_{CC}} &
\ldots \\
\sqrt{S_{CC} S_{AA}} & \sqrt{S_{CC} S_{BB}}  & S_{CC} &
\ldots\\
\ldots & \ldots& \ldots& \ldots
\end{array} \right) \,.
\eqno(3)
$$
In the learning  process according to rule (1),
diagonal elements $S_{AA}$, $S_{BB}$, $S_{CC}\,\ldots$
are determined correctly, while off-diagonal elements appear
to be practically zero; however, they can be corrected
artificially in accordance with the matrix (3). We see
that "teaching to synonyms" is carried out separately,
as it is customary in human practice.

\vspace{2mm}

{\it Homonyms.} From the linguistic point of view, homonyms are
considered as  different words, and this fact is clearly
marked in dictionaries  (usually by Roman numerals, e.g.
{\it like I}, {\it like II}, etc.).
Therefore, no problems arise with homonyms, when a list of
images is composed; they are naturally registered as different
objects. However, in written and conversational text
homonyms are indistinguishable, and the problems arise in the
learning process.

For a given situation, it means that "teaching to homonyms"
should be carried out "by hand": if the computer meets in the
learning text one of the registered homonyms, it should ask the
operator, which of them is implied in the given sentence.
However, such "by hand" stage may be not very long: when
a minimal statistics is
obtained for a correlation
matrix $A_{ij}$, identification of homonyms can be trusted to the
computer. As a rule, homonyms are used in entirely
different contexts, and their associative links are clearly
different.

\vspace{2mm}

{\it Absence of images.}  Our list of images may be somewhat
incomplete: certain images may be absent in it, if no special
words are invented for them: as a rule, it concerns new or not
very wide-spread images. The latter means that an image is not
sufficiently actual for a population and a lot of people have no
idea of it; so we can forgive
a computer, if it does not know such image.

Of course, such images can be introduced in our list "by hand",
if we ascribe them to some groups of words.
Unfortunately, learning also should be produced by hand: the
computer should ask, does the found combination of words
correspond to the given image, or this combination occured in
the sentence accidently.

\vspace{6mm}
\begin{center}
{\bf 5. Conclusion}
\end{center}
\vspace{3mm}

In the previous sections we suggested a possible variant
of solution for problems 1\,--\,5, formulated in Introduction.
Considering the first three problems, we have in mind a linguistic
information contained in the systems like ABBYY Lingvo. Such
systems clearly distinguish  the main meaning of a word (it
is refered there as "the first meaning") and its
secondary meanings; a list of synonyms is also
attached. Of
course, the latter should be tested on reducibility, but an
algorithm for such test is evident from the discussion. The
ABBY Lingvo system contains also
many combinations of words, some of
which correspond to original images; unfortunately, their
separation requires  additional work.

We have suggested also the analog algorithm for learning and
recognition (problems 4,\,5), which should be considered as
preliminary: it can be developed to a practical level
in the course of experiments based on existing programs
for machine translation \cite{4,5}. We hope that a given analysis
suggests a sufficient material for beginning of such experiments,
and realization of  a "sense of humour" in computers may already
occur in the nearest future.

\vspace{6mm}
\begin{center}
{\it Appendix.} {\bf Correlation of irreducible synonyms}
\end{center}
\vspace{3mm}

It is clear from the text, that irreducible
synonyms are described by a model, according to which  the
appearing image is denoted by word $A$ with probability $p_A$, by
word $B$ with probability $p_B$, by word $C$ with probability
$p_C$, etc.  The repeated appearance of the same image in a short
sentence is rather improbable, and a typical situation
corresponds to existence of two coinciding images in
the neighbouring sentences. The probabilities of the
configurations $AA$, $AB,\,\ldots$ are equil to $p_A^2$, $p_A
p_B,\,\ldots$ correspondingly and  associated with a
correlation matrix
$$
  \left ( \begin{array}{cccc}
p_A^2    &  p_A p_B & p_A p_C &
\ldots \\
p_B p_A & p_B^2     &  p_B p_C &
\ldots \\
p_C p_A & p_C p_B  & p_C^2 &
\ldots\\
\ldots & \ldots& \ldots& \ldots
\end{array} \right) \,,
\eqno(4)
$$
which differs by a constant factor from the matrix (3) due to
arbitrariness in normalization of the latter. The diagonal
elements $S_{AA}$, $S_{BB}$, $S_{CC}\,\ldots$ of the
matrix (3) can be considered as known, since they are correctly
determined by the learning rule (1). The off-diagonal elements
can be established from  correspondence of (3) and (4).

\vspace{4mm}


\end{document}